\documentclass[conference]{IEEEtran}
\IEEEoverridecommandlockouts
\usepackage{balance} 

\usepackage{graphicx}
\usepackage{booktabs}
\usepackage{float}
\usepackage{hyperref}
\usepackage{algorithm}
\usepackage{algorithmicx,algpseudocode}
\usepackage{todonotes}
\usepackage{multirow}
\usepackage{dblfloatfix}
\usepackage{algpseudocode}
\usepackage{booktabs}

\def\NoNumber#1{{\def\alglinenumber##1{}\State #1}\addtocounter{ALG@line}{-1}}
\usepackage{subfigure}
\usepackage{multirow,color,amsmath,amssymb}
\usepackage{array} 
\usepackage{tikz,balance}
\usepackage{enumitem,xcolor,booktabs,colortbl}

\newtheorem{remark}{Remark}

\newcommand{\sC}{\mathcal{C}}

\def\BibTeX{{\rm B\kern-.05em{\sc i\kern-.025em b}\kern-.08em
    T\kern-.1667em\lower.7ex\hbox{E}\kern-.125emX}}

\DeclareRobustCommand*{\IEEEauthorrefmark}[1]{%
    \raisebox{0pt}[0pt][0pt]{\textsuperscript{\footnotesize\ensuremath{#1}}}}

\begin{document}

\title{Constrained Reinforcement Learning for \\Dynamic Material Handling
}

\author{\IEEEauthorblockN{Chengpeng Hu\IEEEauthorrefmark{1,2}, Ziming Wang\IEEEauthorrefmark{1,2}, Jialin Liu\IEEEauthorrefmark{2,1}, Junyi Wen\IEEEauthorrefmark{3}, Bifei Mao\IEEEauthorrefmark{3} and Xin Yao\IEEEauthorrefmark{2,1}}
\IEEEauthorblockA{\IEEEauthorrefmark{1}Research Institute of Trustworthy Autonomous Systems (RITAS),\\
Southern University of Science and Technology, Shenzhen, China.\\}
\IEEEauthorblockA{\IEEEauthorrefmark{2}Guangdong Key Laboratory of Brain-inspired Intelligent Computation,
Department of Computer Science and Engineering, \\Southern University of Science and Technology, Shenzhen, China.\\}
\IEEEauthorblockA{\IEEEauthorrefmark{3}Trustworthiness Theory Research Center, Huawei Technologies Co., Ltd Shenzhen, China.\\
hucp2021@mail.sustech.edu.cn, wangzm2021@mail.sustech.edu.cn, liujl@sustech.edu.cn,\\ wenjunyi@huawei.com, maobifei@huawei.com, xiny@sustech.edu.cn}

\thanks{Corresponding author: Jialin Liu (liujl@sustech.edu.cn).}
}

\maketitle

\begin{abstract}
As one of the core parts of flexible manufacturing systems, material handling involves storage and transportation of materials between workstations with automated vehicles. The improvement in material handling can impulse the overall efficiency of the manufacturing system. However, the occurrence of dynamic events during the optimisation of task arrangements poses a challenge that requires adaptability and effectiveness. In this paper, we aim at the scheduling of automated guided vehicles for dynamic material handling. Motivated by some real-world scenarios, unknown new tasks and unexpected vehicle breakdowns are regarded as dynamic events in our problem. We formulate the problem as a constrained Markov decision process which takes into account tardiness and available vehicles as cumulative and instantaneous constraints, respectively. An adaptive constrained reinforcement learning algorithm that combines Lagrangian relaxation and invalid action masking, named RCPOM, is proposed to address the problem with two hybrid constraints. Moreover, a gym-like dynamic material handling simulator, named DMH-GYM, is developed and equipped with diverse problem instances, which can be used as benchmarks for dynamic material handling. Experimental results on the problem instances demonstrate the outstanding performance of our proposed approach compared with eight state-of-the-art constrained and non-constrained reinforcement learning algorithms, and widely used dispatching rules for material handling. 
\end{abstract}




\maketitle 


\section{Introduction}
\begin{figure}[t]
    \centering
    \includegraphics[width=0.85\linewidth]{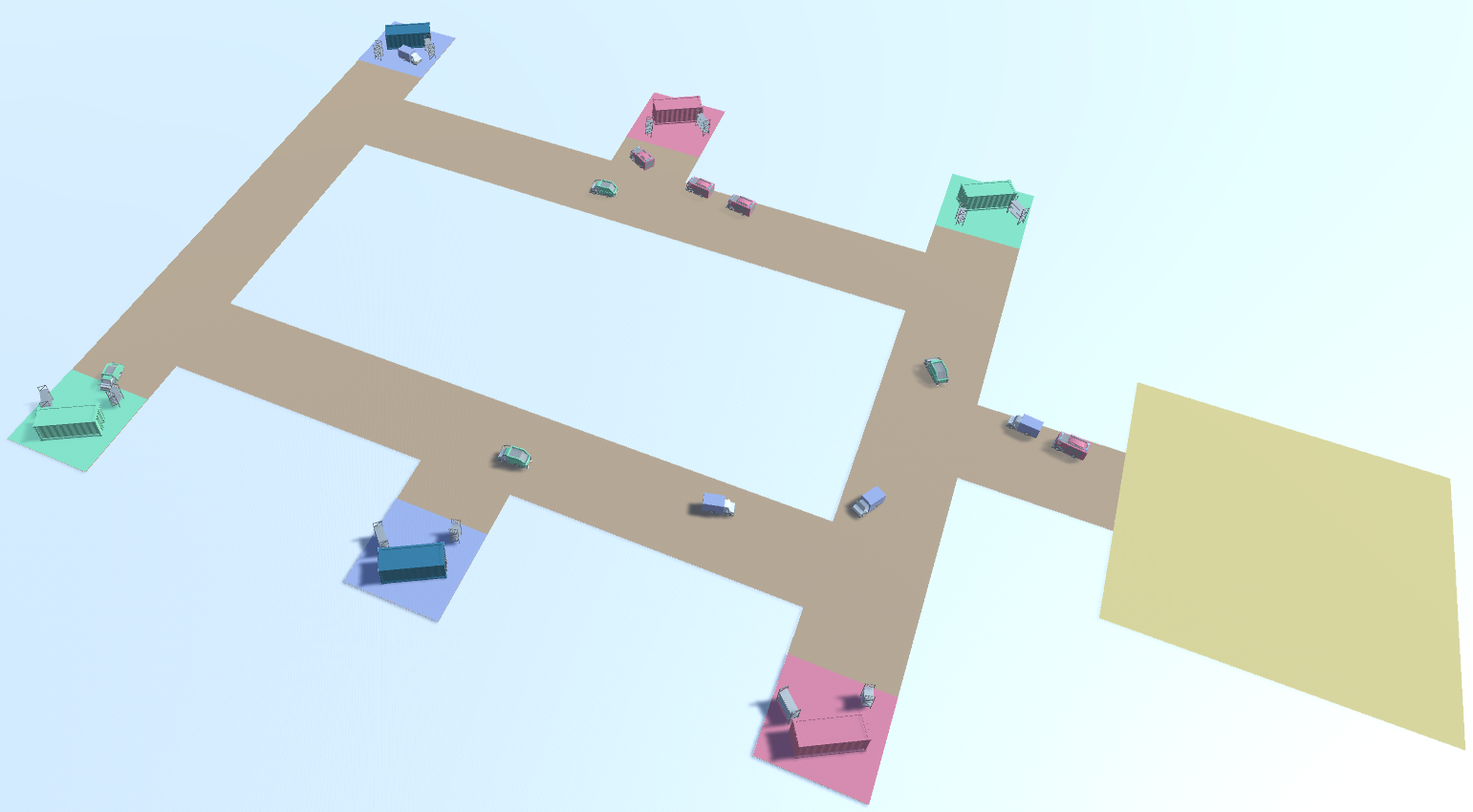}
    \caption{Simulation of material handling.}
    \label{fig:mhsim}
\end{figure}

Material handling can be widely found in manufacturing, warehouses and other logistic scenarios. It aims to transport some goods from their storage locations to some delivery sites. 
With the help of automated guided vehicles (AGV), tasks and jobs can be accomplished fast and automatically. An example of material handling is shown in Fig. \ref{fig:mhsim}. In real-world flexible manufacturing, AGV scheduling plans usually need to be changed due to some dynamics such as newly arrived tasks, due time change, as well as vehicle and site breakdowns~\cite{kaplanouglu2015multi}. These dynamics pose a serious challenge to vehicle scheduling, called dynamic material handling (DMH).

Dispatching rules~\cite{raghu1993efficient,sabuncuoglu1998study} are classic and common methods for DMH. Although easy to implement, they suffer from some complex situations, which usually lead to suboptimal performance and are hard to be improved further. Considering such issues, search-based methods come to prominence. 
For example, a hybrid genetic algorithm and ant colony optimisation is proposed for intelligent manufacturing workshop~\cite{xu2019dynamic}.
However, the search process is often time-consuming. The aforementioned methods are less suitable in real-world scenarios when fast response and adaptability are needed.

Recently, some work has leveraged reinforcement learning (RL) to DMH~\cite{xue2018reinforcement,hu2020deep}. The problem is formulated as a Markov decision process (MDP) and the reward function is constructed manually according to makespan and travel distance~\cite{xue2018reinforcement,hu2020deep}.
Trained parameterised policies schedule the vehicles in real time when new tasks arrive.
However, more complicated scenarios involving vehicle breakdowns or some vital problem constraints, such as task delay, are not considered while optimising the policy~\cite{xue2018reinforcement,hu2020deep}. It is impossible to overlook these scenarios when dealing with real-world problems. For example, manufacturing accidents may happen if an emerged task is assigned to a broken vehicle, which leads to a task delay or this task will never be completed.



Motivated by real-world scenarios in the flexible manufacturing system, we consider a DMH problem with multiple AGVs and two dynamic events, namely newly arrived tasks and vehicle breakdowns during the handling process. To tackle the dynamics and constraints, we formulate the problem as a constrained Markov decision process (CMDP), considering the tardiness of tasks as a cumulative constraint. In addition, the status of vehicles including ``idle'', ``working'' and ``broken'' caused by vehicle breakdowns is considered as an instantaneous constraint which determines if a vehicle can be assigned with a task or not. 
A constrained reinforcement learning (CRL) approach is proposed to perform real-time scheduling considering two widely used performance indicators, makespan and tardiness in manufacturing. The proposed approach, named RCPOM, incorporates Lagrangian relaxation and invalid action masking to handle both cumulative constraint and instantaneous constraint.
Moreover, due to the lack of free simulators and instance datasets for DMH, we develop an OpenAI's gym-like~\cite{1606.01540} simulator and provide diverse problem instances as a benchmark\footnote{Codes available at \url{https://github.com/HcPlu/DMH-GYM}}. 

The main contributions of our work are summarised as follows.
\begin{itemize}
    \item We formulate a dynamic material handling problem considering multiple vehicles, newly arrived tasks and vehicle breakdowns as a CMDP with the consideration of both cumulative and instantaneous constraints. 
    \item Extending OpenAI's gym, we develop an open-source simulator, named DMH-GYM, and provide a dataset comprised of diverse instances for researching the dynamic material handling problem.
    \item Considering the occurrence of newly arrived tasks and vehicle breakdowns as dynamic events, a CRL algorithm combining a Lagrangian relaxation method and invalid action masking technique, named RCPOM, is proposed to address this problem.
    \item Extensive experiments show the adaptability and effectiveness of the proposed method for solving this problem, compared with eight state-of-the-art algorithms, including CRL agents, RL agents and several dispatching rules.
\end{itemize}
\def\nouse{
The paper is organised as follows. Section \ref{sec:back} introduces the background of dynamic material handling. Section \ref{sec:dmh} describes the problem and developed simulator, DMH-GYM. 
We model the problem as a CMDP in Section \ref{sec:crl}. Section \ref{sec:crl} also presents the details of the proposed approach that combines Lagrangian relaxation and invalid action masking. Then, in Section \ref{sec:exp}, we validate our proposed approach in training and unseen test DMH instances compared with several state-of-the-art CRL agents, RL agents and common dispatching rules. Section \ref{sec:conclusion} concludes.
}



\section{Background}
\label{sec:back}

AGVs are widely used for material handling systems~\cite{qiu2002scheduling}, meeting the need for the high requirement of automation. Scheduling and routing are usually regarded as two splitting parts of material handling. The former refers to the task dispatching considered in the paper. The latter aims to find feasible routing paths for AGVs to transfer loads. Pathfinding is not the main focus of this paper, since fixed paths are commonly found in many real-world cases~\cite{qiu2015heterogeneous,singh2022matheuristic,dang2021scheduling}.

Scheduling of DMH~\cite{ouelhadj2009survey} refers to the problem with some real-time dynamic events, such as newly arrived tasks, due time change and vehicle breakdowns~\cite{kaplanouglu2015multi}, which are commonly found in modern flexible manufacturing systems and warehouses. Traditional dispatching rules~\cite{sabuncuoglu1998study}, e.g., first come first serve (FCFS), earliest due date first (EDD), and nearest vehicle first (NVF), are widely used to solve the dynamic optimisation problem due to their simplicity. Although reasonable solutions may be found by these rules, they hardly get further improvement. Dispatching rules are usually constructed with some specific considerations, thus few certain single rules work well in general cases~\cite{blackstone1982state}. Mix rule policy that combines multiple dispatching rules improves the performance of a single one~\cite{chen2011multiple}. New AGVs and jobs are considered dynamic events and solved with the multi-agent negotiation method~\cite{sahin2017multi}. An adaptive mixed integer programming model triggered by some events is applied to solve the single AGV scheduling problem with time windows~\cite{liu2018dynamic}. Nevertheless, these methods are usually limited for poor adaptability and require domain knowledge.

Search-based methods can also be applied to DMH. Since priory conditions such as the number of tasks are not determined apriori, these methods seek to decompose the dynamic process as some static sub-problems. Genetic algorithm~\cite{chryssolouris2001dynamic} is applied for DMH. When new tasks arrive or the status of machines changes, the algorithm reschedules the plan. Analytic hierarchy process based optimisation~\cite{zhang2015optimization} combines task sets and then uses mixed attributes for dispatching. NSGA-II is used by Wang et al.~\cite{wang2020proactive} to achieve proactive dispatching considering distance and energy consumption at the same time. However, these search-based methods usually suffer from long consumption time ~\cite{chryssolouris2001dynamic}, and thus are not able to cater to the requirements of fast response and adaptability in real-world scenarios.

Reinforcement learning, which performs well in real-time decision-making problems~\cite{haarnoja2018soft}, has been applied to optimise real-world DMH problems.
A Q$-\lambda$ algorithm is proposed by Chen et al.~\cite{chen2015reinforcement} to address a multiple-load carrier scheduling problem in a general assembly line.
Xue et al. \cite{xue2018reinforcement} consider a non-stationary case that AGVs' actions may affect each other. 
Kardos et al. \cite{kardos2021dynamic} use Q-learning to schedule tasks dynamically with real-time information.
Hu et al. \cite{hu2020deep} improve the mix rule policy~\cite{chen2011multiple} with a deep reinforcement learning algorithm to minimise the makespan and delay ratio. Multiple AGVs are scheduled by the algorithm in real time. \cite{govindaiah2021applying} applies reinforcement learning with a weight sum reward function to material handling. 
Although the algorithms of the above work perform promisingly, constraints in real-world applications are not considered in their problems. 

Constrained reinforcement learning (CRL) follows the principles of the constrained Markov decision process (CMDP) that maximises the long-term expected reward while respecting constraints~\cite{altman1999constrained}. It has been applied to solving problems such as robot controlling~\cite{achiam2017constrained,tessler2018reward} and resource allocation~\cite{bhatia2019resource,liu2020constrained}, however, to the best of our knowledge, it has never been considered in solving DMH problems.


\section{Dynamic Material Handling with multiple vehicles}
\label{sec:dmh}

This section describes our DMH, considering newly arrived tasks and vehicle breakdowns. DMH-GYM, the simulator developed in this work is also presented.

\subsection{Problem description}

In our DMH problem with multiple vehicles, transporting tasks can be raised over time dynamically. These newly arrived tasks, called \emph{unassigned tasks}, need to be assigned to AGVs with a policy $\pi$.
The problem is formed as a graph $G(\mathcal{L},\mathcal{T})$ where $\mathcal{L}$ and $\mathcal{T}$ denote the sets of sites and paths, respectively. Three kinds of sites, namely \emph{pickup points}, \emph{delivery points} and \emph{parking positions} are located in graph $G(\mathcal{L},\mathcal{T})$. The set of parking positions is denoted as $\mathcal{K}$. The total number of tasks that need to be completed is unknown in advance. All unassigned tasks will be stored in a staging list $\mathcal{U}$ temporarily. A \emph{task} $u=\langle s,e,\tau,o \rangle \in \mathcal{U}$ is determined by its \emph{pickup point} $u.s$, \emph{delivery point} $u.e$, \emph{arrival time} $u.\tau$ and \emph{expiry time} $u.o$, where $u.s,u.e \in\mathcal{L}$.

A fleet of AGVs $\mathcal{V}$ works in the system to serve tasks. A task can only be picked by one and only one AGV at once. 
Each AGV $v= \langle vl,rp,pl,\psi \rangle$ $\in \mathcal{V}$ is denoted by its \emph{velocity} $v.vl$, \emph{repairing time} $v.rp$, \emph{parking position} $v.pl$ and \emph{status} $v.\psi$. The location $v.pl$ denotes the initial parking location of AGV $v$, where $v.pl \in \mathcal{K}$. AGVs keep serving the current assigned task $v.u$ during the process. If there is no task to be performed, the AGV stays at its current place until a new task is assigned. All finished tasks by an AGV $v$ will be recorded in the historical task list $HL(v)$. 
A starting task $u_0$ will first be added to the historical list, where $u_0$ denotes that an AGV $v$ leaves its parking site to the pickup point of the first assigned task $u_1$. Three statuses of AGV are denoted as set $\Psi =\{I,W,B\}$ representing \emph{Idle}, \emph{Working} and \emph{Broken}, respectively. An AGV can break down sometimes. In this broken status, the broken AGV $v$ will stop in place and release its current task to the task pool. After repairing in time $v.rp$, AGV can be available to accept tasks. 

While the system is running, a newly arrived task $u$ can be assigned to an available AGV if and only if $v.\psi=I$.
The decision policy $\pi$ assigns the task. 
 Makespan and tardiness are selected as the optimising objectives of the problem, especially the tardiness is regarded as a constraint. Makespan is the maximal time cost of AGVs for finishing all the assigned tasks using Eq. \eqref{eq:makespan}. Tardiness denotes the task delay in the completion of tasks, as formalised in Eq. \eqref{eq:tardiness}.
\begin{eqnarray}
&&\hspace{-12mm}F_{m}(\pi) = \max_{v \in \mathcal{V}^{\pi}} FT(u^v_{|HL(v)|},v),\label{eq:makespan}\\
&&\hspace{-12mm}F_{t}(\pi) = \frac{1}{m}\sum_v^{\mathcal{V}^{\pi}}\sum_{i}^{|HL(v)|}\max(FT(u^v_{i},v)-u^v_{i}.o-u^v_{i}.\tau,0), \label{eq:tardiness}
\end{eqnarray}
where $HL(v)$ refers to the historical list of tasks completed by $v$ and $v.pl \in \mathcal{P}$. The time cost of waiting to handle material is ignored at pickup points and delivery points for AGVs as they are often constant.



Notations for our problem are summarised as follows.

\noindent$|\cdot|$: Size of a given set or list.\\
\noindent$\mathcal{V}$: A fleet of AGVs.\\
\noindent$n$: Number of AGVs, i.e., $n=|\mathcal{V}|$.\\
\noindent$m$: Number of tasks that needs to be accomplished.\\
\noindent$\mathcal{L}$:  Set of pickup points, delivery points, and parking lots. \\
\noindent$\mathcal{T}$: Paths set that connect sites. \\
\noindent$\mathcal{K}$: Parking set, $\mathcal{K} \subset \mathcal{L}$.\\
\noindent$\mathcal{U}$: Staging list of tasks.\\
\noindent$u$: Task $u \in \mathcal{U}$ is denoted as a tuple $\langle s,e,\tau,o\rangle$ that refers to pickup point $s$, delivery point $e$, arrival time $\tau$ and expiry time $o$.\\
\noindent$u.s$: Pickup point of a given task $u$, $u.s \in \mathcal{L}$.\\
\noindent$u.e$: Delivery point of a given task $u$, $u.e \in \mathcal{L}$.\\
\noindent$u.\tau$: Arrival time of a given task $u$.\\
\noindent$u.o$: Expiry time of a given task $u$.\\
\noindent$u_0$: Start task that manipulates AGV to leave its parking lot.\\
\noindent$v$: AGV $v$ is denoted as tuple $\langle vl,rp,pl,\psi \rangle$ that refers to its velocity $vl$, repair time $rp$, parking location $pl$ and status $\psi$.\\
\noindent$v.vl$: Velocity of a given AGV $v$. \\
\noindent$v.rp$: Repairing time of a given AGV $v$. \\
\noindent$v.pl$: Parking position of a given AGV $v$, $v.pl \in \mathcal{K}$.\\
\noindent$v.\psi$: Status of a given AGV $v$. Three statuses are defined as \emph{Idle}, \emph{Working} and \emph{Broken} and represented by set $\Psi =\{I,W,B\}$.\\
\noindent$v.u$: Current assigned task of a given AGV $v$.\\
\noindent$HL(v)$: List of completed tasks of a given AGV $v$.\\
\noindent$FT(u,v)$: Finish time of a task $u$ by an AGV $v$.\\
\noindent$\pi$: Decision policy for task assignments.\\


\subsection{DMH-GYM}
\begin{figure}[htbp]
    \centering
    \includegraphics[width=0.73\linewidth]{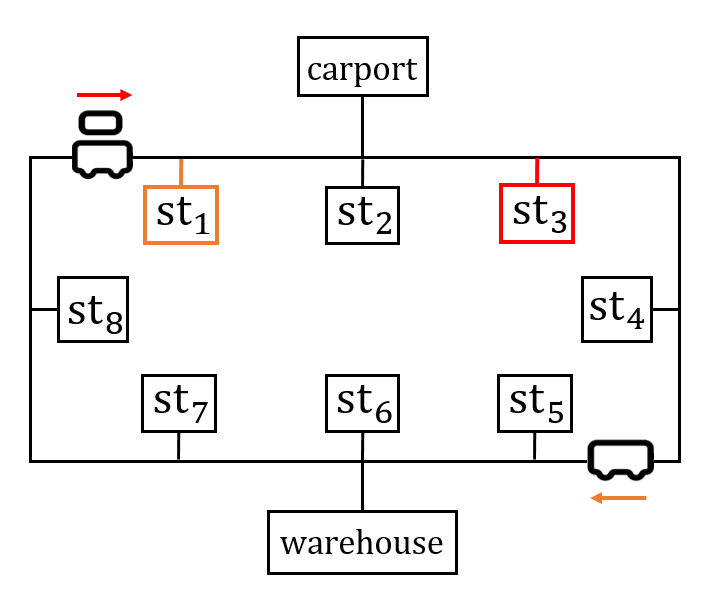}
    \caption{Layout of DMH-GYM.}
    \label{fig:envlayout}
\end{figure}

To our best knowledge, no existing work has studied this problem and there is no existing open-source simulator of related problems.
To study the problem, we develop a simulator that is compatible with OpenAI's gym~\cite{1606.01540}, named DMH-GYM. 
Diverse instances are also provided that show various performances on different dispatching rules to fully evaluate the given algorithms.
The layout of the shop floor is shown in Fig. \ref{fig:envlayout}. Stations from st1 to st8 are the workstations on the floor which can be pickup points and delivery points. The warehouse can only be a delivery point. All AGVs depart from the carport, i.e. initial parking position, to accomplish tasks that are generated randomly. The start site of a task can only be a workstation, while its end site can be a workstation or warehouse. All AGVs move on preset paths with the same velocity. And they are guided by an $A^*$ algorithm for path finding while performing tasks.
The distance between stations is calculated according to their coordinates and path. For example, the distance between st8 (0,45) and st1 (20,70) in Fig. \ref{fig:envlayout} is 45 as a vehicle should pass the coordinate of the corner point (0,70).
Collisions and traffic congestion are not considered in this paper for simplification.

\section{Constrained Reinforcement Learning}
\label{sec:crl}
To meet the needs of real-world applications, we extend the DMH problem as a CMDP by considering the tardiness and the constraint of vehicle availability.
\subsection{Modeling DMH as a CMDP}
The CMDP is defined as a tuple $(\mathcal{S},\mathcal{A},\mathcal{R},\mathcal{C},\mathcal{P},\gamma)$, where $\mathcal{S}$ is state space, $\mathcal{A}$ is action space, $\mathcal{R}$ represents the reward function, $\mathcal{P}$ is the transition probability function and $\gamma$ is the discount factor. A cumulative constraint $\mathcal{C} = g(c(s_0,a_0,s_{1}),\dots,c(s_{t},a_{t},s_{t+1}))$ that consists of per-step cost $c(s,a,s')$ can be restricted by a corresponding constraint threshold $\epsilon$. $g$ can be any of the functions for constraints such as average and sum. $J_{\mathcal{C}}^{\pi}$ denotes the expectation of the cumulative constraint.
A policy optimised by parameters $\theta$ is defined as $\pi_{\theta}$ which determines the probability to take an action $a_t$ at state $s_t$ with $\pi(a_t|s_t)$ at time $t$. The goal of the problem is to maximise the long-term expected reward with optimised policy $\pi_{\theta}$ while satisfying the constraints, as formulated below.
\begin{eqnarray}
&&\max_{\theta}~ \mathbb{E}_{\pi_{\theta}}[\sum \limits_{t=0}^{\infty}\gamma^t\mathcal{R}(s_t,a_t,s_{t+1})]  \\
s.t. \noindent&&  J_{\mathcal{C}}^{\pi_{\theta}} = \mathbb{E}_{\tau \sim \pi_{\theta}}[\sC] \leq \epsilon. 
\end{eqnarray}

\subsubsection{State}
At each decision time $t$, we consider the current state of the whole system consisting of tasks and AGVs, denoted as $S_t=\rho( \mathcal{U}_t, \mathcal{V}_t)$,
where $\mathcal{U}_t$ is the set of unassigned tasks and $\mathcal{V}_t$ represents information of all AGVs at time $t$.
We encode the information of the system at decision time $t$ with a feature extract function $\rho(\mathcal{U}_t, \mathcal{V}_t)$ as structure representation. Specifically, a state is mapped into a vector by $\rho$ consisting of the number of unassigned tasks, tasks' remaining time, waiting time and distance between pickup and delivery point as well as statuses of vehicles and the minimal time from the current position to pickup point then delivery point.

\subsubsection{Action}\label{sec:action}
The goal of dynamic material handling is to execute all tasks, thus the discrete time step refers to one single, independent task assignment. Different from the regular MDP like in games, the number of steps in our problem is usually fixed. In other words, the length of an episode is decided by the number of tasks to be assigned.
In our CMDP, making an action is to choose a suitable dispatching rule and a corresponding vehicle. A hybrid action space $\mathcal{A}_t=\mathcal{D} \times \mathcal{V}_t$ is considered for the CMDP, where $\mathcal{D}$ and $\mathcal{V}_t$ are defined as dispatching rule space and AGV space, respectively. Four dispatching rules, including first come first served (FCFS), shortest travel distance (STD), earliest due date first (EDD) and nearest vehicle first (NVF), form the dispatching rule space $\mathcal{D}$.
Those dispatching rules are formulated~\cite{sabuncuoglu1998study} as follows:
\begin{itemize}
    \item FCFS: $\arg\min \limits_{u\in \mathcal{U}} u.\tau$;
    \item EDD: $\arg\min \limits_{u\in \mathcal{U}} (u.o-u.\tau$);
    \item NVF: $\arg\min \limits_{u\in \mathcal{U}} d(v.c,u.s)$;
    \item STD: $\arg\min \limits_{u\in \mathcal{U}} d(v.c,u.s)+d(u.s,u.e)$,
\end{itemize}
where $v.s$ is the current position of an AGV $v$ and $d(p,p')$ determines the distance between $p$ and $p'$. Action $a_t = \langle d_t,v_t \rangle \in \mathcal{A}_t$ at time $t$ determines a single task assignment $a_t$ that task $u_t$ is assigned to AGV $v_t$ by dispatching rule $d_t$. With the action space $\mathcal{A}_t$, the policy can decide the rule and AGV at the same time, which breaks the tie of the multiple vehicles case.


\subsubsection{Constraints}
A cumulative constraint and an instantaneous constraint are both considered in the CMDP. We treat the tardiness as a cumulative constraint $J_{\mathcal{C}}$, formulated in Eq. \eqref{eq:cm}. The task assignment constraint of vehicle availability is regarded as an instantaneous constraint that only vehicles with status \emph{Idle} are considered as available, denoted as Eq. \eqref{eq:im}.


\begin{eqnarray}
&&J_{\mathcal{C}}^{\pi_\theta} = \mathbb{E}_{\pi_\theta}(F_t(\pi_\theta)) \leq \epsilon, \label{eq:cm} \\
    && v.\psi=I. \label{eq:im} 
\end{eqnarray}

\subsubsection{Reward function}
The negative makespan returned at the last timestep is set as a reward. $\tau$ is denoted as a trajectory $ (s_0,a_0,s_1,a_1,\dots,s_t,a_t,s_{t+1},\dots)$ sampled from current policy $\pi$. The per-step reward function is defined as follows.
\begin{eqnarray}
\mathcal{R}(s_t,a_t,s_{t+1})=
\left\{
             \begin{array}{ll}
             -F_{m}(\pi),& \text{ if terminates},  \\
             0, &\text{ otherwise}. 
             \end{array}
\right.
\label{eq:rewardf}
\end{eqnarray}

\subsection{Constraint handling}
We combine invalid action masking and the reward constrained policy optimisation (RCPO)~\cite{tessler2018reward}, a Lagrangian relaxation approach, to handle the instantaneous and cumulative constraints at the same time.
\begin{figure}[t]
    \centering
    \includegraphics[width=0.9\linewidth]{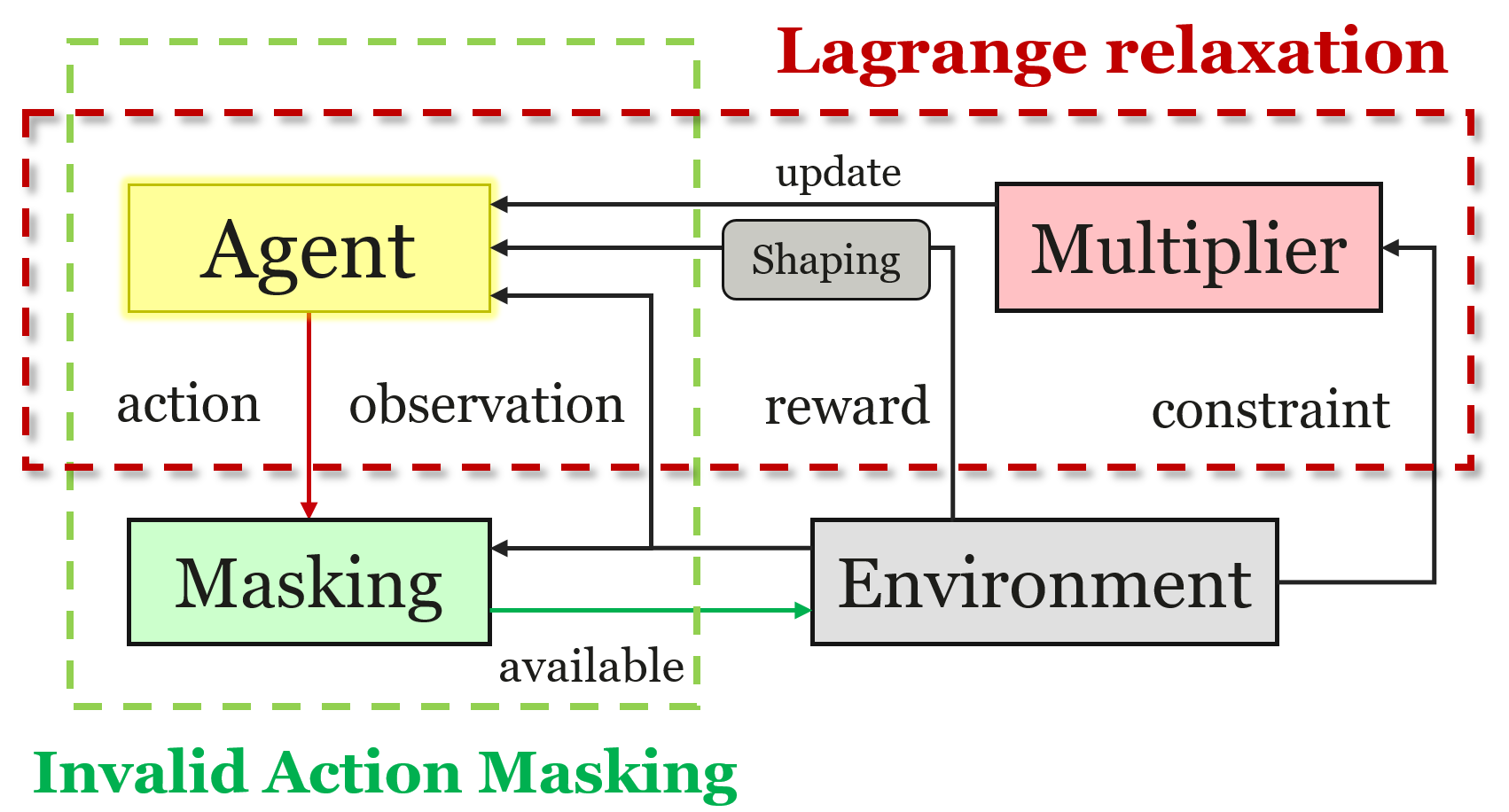}
    \caption{Reward constrained policy optimisation with masking.}
    \label{fig:framework}
\end{figure}
\subsubsection{Feasible action selection via masking}
\begin{algorithm}[b]
\caption{Invalid action masking.}
\label{alg:masking}
\begin{algorithmic}[1] 
\Require policy $\pi_{\theta}$, state $s$
\Ensure action $a$
\State Logits $l\leftarrow \pi_{\theta}(\cdot|s)$
\State Compute $l'$ by replacing $\pi_{\theta}(a'|s)$ with $-\infty$ where $a'$ is an invalid action
\State $\pi'_{\theta}(\cdot|s) \leftarrow$ softmax($l'$)
\State Sample $a \sim \pi'_{\theta}(\cdot|s)$

\end{algorithmic}
\end{algorithm}
We can indeed determine the available actions according to the current state. For example, in our problem, the dimension of the current action space can be 4, 8, ..., which depends on the available vehicles at any time step. However, since the dimension of the input and output of neural networks used are usually fixed, it is hard to address the variable action space directly.

To handle the instantaneous constraint, we adapt the invalid action masking technique that has been widely applied in games~\cite{ye2020mastering,huang2020closer}. The action space is then compressed into $\mathcal{A}_t = \mathcal{D} \times \mathcal{AV}_t$ with the masking, where $\mathcal{AV}_t$ is the set of available AGVs.
Logits corresponding to all invalid actions will be replaced with a large negative number, and then the final action will be resampled according to the new probability distribution. Masking helps the policy to choose valid actions and avoid some dangerous situations like getting stuck somewhere.
Pseudo code is shown in Alg. \ref{alg:masking}.

It is notable that, to the best of our knowledge, no existing work has applied the masking technique to handle the instantaneous constraint in DMH. The closest work is presented by \cite{hu2020deep}, which applies the DQN~\cite{mnih2015human} with a designed reward function.
According to \cite{hu2020deep}, the DQN agent selects the action with maximal Q value and repeats this step until a feasible action is selected, i.e., assigning a task to an available vehicle. However, when applying the method of \cite{hu2020deep} to our problem, the agent will keep being trapped in this step if the selected action is infeasible. Even though keeping resampling the action, the same feasible action will always be selected by this deterministic policy at a certain state.

\subsubsection{Reward constrained policy optimisation with masking}
Pseudo code of our proposed method is shown in Alg. \ref{alg:rcpom}. We combine RCPO~\cite{tessler2018reward} and invalid action masking, named RCPOM, to handle the hybrid constraints at the same time, seeing Fig. \ref{fig:framework}. RCPO is a Lagrangian relaxation type method based on Actor-Critic framework~\cite{mnih2016asynchronous} which converts a constrained problem into an unconstrained one by introducing a multiplier $\lambda$, seeing Eq. \eqref{eq:unP},
\begin{eqnarray}
\min_{\lambda}\max_{\theta}[\mathcal{R}^\pi-\lambda (J_\mathcal{C}^{\pi} - \epsilon)],\label{eq:unP}
\end{eqnarray}
where $\lambda>0$. The reward function is reshaped with Lagrangian multiplier $\lambda$ as follows:
\begin{equation}\label{eq:reshapeR}
    \hat{\mathcal{R}}(s,a,s',\lambda) = \mathcal{R}(s,a,s')-\lambda c(s,a,s').
\end{equation}

\def\nouse{
Thus, the value function can be formulated as Eq. \eqref{eq:V}.
\begin{equation}
\begin{aligned}
V^{\pi}&=\mathbb{E}_{\pi}[\sum \limits_{t=0}^{\infty}\gamma^t\hat{\mathcal{R}}(s_t,a_t,s_{t+1},\lambda)|s_0=s] \\
&=\mathbb{E}_{\pi}[\sum \limits_{t=0}^{\infty}\gamma^t(\mathcal{R}(s_t,a_t,s_{t+1})-\lambda c(s_t,a_t,s_{t+1}))|s_0=s]\\
&=V^{\pi}_{\mathcal{R}}-\lambda V^{\pi}_C,
    \label{eq:V}
\end{aligned}
\end{equation}
where $V^{\pi}_{\mathcal{R}}$ and $\lambda V^{\pi}_C$ represent value functions of reward and constraint, respectively.
}
In a slower time scale than the update of the actor and critic, the multiplier is updated with the collected constraint values according to Eq. \eqref{eq:rcpou}.
\begin{eqnarray}
    \lambda = \max\{\lambda+\eta(J_\mathcal{C}^{\pi}-\epsilon),0\}, \label{eq:rcpou}
\end{eqnarray}
where $\eta$ is the learning rate of the multiplier.

Specifically, we implement RCPOM with soft actor critic (SAC) because of its remarkable performance~\cite{haarnoja2018soft}. SAC is a maximum entropy reinforcement learning that maximises cumulative reward as well as the expected entropy of the policy. The objective function is adapted as follows:
\begin{eqnarray}
    J(\pi_{\theta}) = \mathbb{E}_{\pi_{\theta}}[\sum \limits_{t=0}^{\infty}\gamma^t(\mathcal{R}(s_t,a_t,s_{t+1})+\alpha \mathcal{H}(\pi(\cdot|s_t)))],
\end{eqnarray}
where the temperature $\alpha$ decides how stochastic the policy is, which encourages exploration.

\begin{table*}[h]
    \centering

        \caption{Time consumed for making a decision averaged over 2000 trials.}
                \resizebox{\textwidth}{!}{
          \setlength{\tabcolsep}{4pt}
    \begin{tabular}{c|c|c|c|c|c|c|c|c|c|c|c|c|c}
    \toprule
         & RCPOM & RCPOM-NS &IPO &  L-SAC & L-PPO & SAC & PPO & FCFS & EDD & NVF & STD & Random & Hu et al. \cite{hu2020deep} \\
         \midrule
         Time (ms) & 2.143& 2.112& 2.038& 2.74& 2.514& 2.79 & 2.626 & 0.0169 & 0.0199 & 0.0239 & 0.0354 & 0.0259 & Timeout\\
        \bottomrule
    \end{tabular}
}
    \label{tab:ptime}
\end{table*}
\begin{algorithm}[t]
\caption{RCPO with masking (RCPOM).}
\label{alg:rcpom}
\begin{algorithmic}[1] 
\Require Epoch $N$, Lagrange multiplier $\lambda$, learning rate of Lagrange multiplier $\eta$, temperature parameter $\alpha$
\Ensure Policy $\pi_\theta$
\State Initialise an experience replay buffer $\mathcal{B}_R$
\State Initialise policy $\pi_{\theta}$ and two critics $\hat{Q}_{\psi_1}$ and $\hat{Q}_{\psi_2}$ 
\For{ $n=1$ to $N$}
\For{$t=0,1,\dots$}
    \State $a_t' \leftarrow $ \textbf{Masking}($\pi_{\theta},s_t$) \Comment{Alg. \ref{alg:masking}}
    \State Get $s_{t+1},r_t,c_t$ by executing $a'_t$
     \State Store $\langle s_t,a_t',s_{t+1},r_t,c_t\rangle$ into $\mathcal{B}_R$
    \State Randomly sample a minibatch $B_R$ of transitions $\mathcal{T}= \langle s,a,s',r,c\rangle$ from $\mathcal{B}_R$
    \State Compute $y \leftarrow r-\lambda c+\gamma (\min\limits_{j=1,2}\hat{Q}_{\psi_j}(s',\tilde{a}')-\alpha \log{\pi_\theta(\tilde{a}'|s')})$, where $\tilde{a}'\sim \pi_{\theta}(\cdot|s') $
    \State Update critic with 
    \NoNumber{$\nabla_{\psi_j}\frac{1}{|B_R|}\sum\limits_{\mathcal{T} \in B_R}^{}(y-\hat{Q}_{\psi_j}(s,a))^2$ for $j=1,2$}
    \State Update actor with 
    \NoNumber{$\nabla_{\theta}\frac{1}{|B_R|}\sum\limits_{\mathcal{T} \in B_R}(\min\limits_{j=1,2}\hat{Q}_{\psi_j}(s,\tilde{a}_{\theta})-\alpha \log\pi_\theta(\tilde{a}_{\theta}|s))$, $\tilde{a}_{\theta}$ is sampled from $\pi_\theta(\cdot|s)$ via reparametrisation trick}
    \State Apply soft update on target networks
    \EndFor
    \State $\lambda \gets \max (\lambda+\eta(J_\mathcal{C}^{\pi}-\epsilon),0)$ \Comment{Eq. \eqref{eq:rcpou}}
\EndFor
\end{algorithmic}
\end{algorithm}

\subsubsection{Invariant reward shaping} The raw reward function is reshaped with linear transformation for more positive feedback and better estimation of the value function, seeing Eq. \eqref{eq:reshapeIR}
\begin{equation}\label{eq:reshapeIR}
    \tilde{\mathcal{R}}(s,a,s',\lambda) = \beta \mathcal{R}(s,a,s')+b,
\end{equation}
where $\beta>0$ and $b$ is a positive constant number.
It is easy to guarantee policy invariance under this reshape~\cite{sutton2018reinforcement}, declared in Remark \ref{remark:rs}. 
\begin{remark}
Given a CMDP $(\mathcal{S},\mathcal{A},\mathcal{R},\mathcal{C},\mathcal{P},\gamma)$, the optimal policy keeps invariant with linear transformation in Eq. \ref{eq:reshapeIR}, where $\beta>0$ and $b \in \mathbb{R}$:
\begin{equation*}
    \forall s\in \mathcal{S}, V^*(s) = \mathop{\arg\max}_{\pi} V^{\pi}(s)= \mathop{\arg\max}_{\pi} \beta V^{\pi}(s)+\frac{b}{1-\gamma}.
\end{equation*}

\label{remark:rs}
\end{remark}

\section{Experiments}

\label{sec:exp}
To verify our proposed method, numerous experiments are conducted and discussed in this section. 

The proposed method denoted as ``RCPOM'' is compared with five groups of methods: (i) an advanced constrained reinforcement learning agent Interior-point Policy Optimisation (IPO)~\cite{liu2020ipo}, (ii) state-of-the-art reinforcement learning agents including proximal policy optimization (PPO)~\cite{schulman2017proximal} and soft actor critic (SAC)~\cite{haarnoja2018soft}, (iii) SAC and PPO with fixed Lagrangian multiplier named as ``L-SAC'' and ``L-PPO'', and (iv) commonly used dispatching rules including FCFS, STD, EDD and NVF as presented in Section \ref{sec:action}. (v) To validate the invariant reward shaping, an ablation study is also performed. The RCPOM agent without the invariant reward shaping denoted as ``RCPOM-NS'', is compared. A random agent is also compared.

The simulator DMH-GYM and 16 problem instances are used in our experiments, where instance01-08 are training instances and instance09-16 are unseen during training. Trials on the training instances and unseen instances verify the effectiveness and adaptability of our proposed method. 





\subsection{Settings}

We apply the invalid action masking technique to all the learning agents to ensure safety when assigning tasks since the instantaneous constraints should be satisfied per time step. All learning agents except ``RCPOM-NS'' are equipped with the reward shaping as a fair comparison with the proposed method.
To ensure instantaneous constraint satisfaction and explore more possible plans, we adapt the dispatching rules. 
Dispatching rules consider current feasible task assignments and shuffle these possible task assignments when the case of multiple available vehicles is met.

All learning agents are adapted based on the implementation of Tianshou framework~\cite{weng2021tianshou}~\footnote{https://github.com/thu-ml/tianshou}. The network structure is formed by two hidden fully connected layers $128 \times 128$.
$\alpha$ is 0.1. $\gamma$ is 0.97. The initial multiplier $\lambda$ and its learning rate are set as 0.001 and 0.0001, respectively.
The constraint threshold $\epsilon$ is set as 50. Reward shaping parameters $\beta$ and $b$ are set as 1 and 2000 in terms of dispatching rules, respectively.
All learning agents are trained for 5e5 steps on an Intel Xeon Gold 6240 CPU and four TITAN RTX GPUs. The best policies during training are selected. All dispatching policies and learning agents are tested 30 times on the instances independently. Parameter values are either set following previous studies~\cite{tessler2018reward,weng2021tianshou} or arbitrarily chosen.


\subsection{Experimental result}

Tab. \ref{tab:result} and Tab. \ref{tab:unseenResult} present the results on the training and unseen instances, respectively. The average time consumption per task assignment is demonstrated in Tab. \ref{tab:ptime}.

It is obvious that the random agent performs the worst. In terms of dispatching rules, EDD shows its ability to optimise the time-based metric, namely tardiness. In all instances, the tardiness of EDD is almost under the tardiness constraint threshold. For Instance01-Instance03, EDD gets the lowest tardiness 33.9, 29.6 and 26.5 compared with other policies, respectively. FCFS is also a time-related rule that always assigns the tasks that arrive first. It only performs better than the other three rules in Instance07 for its low makespan and tardiness. Two other distance-related rules, NVF and STD that optimise the travel distance can achieve good results on makespan. For example, STD has the smallest makespan 1883.5 in Instance08 among all the policies. However, both rules fail in terms of tardiness, which is shown in Tab. \ref{tab:result} for their relatively high constraint value.


Learning agents usually show better performance on makespan than dispatching rules. In Instance01, the SAC agent gets the best makespan of 1798.4. The proposed RCPOM performs the best among all the policies in terms of makespan on most of the training instances except Instance01, Instance03 and Instance08. Although STD achieves 1883.5 on Instance08, RCPOM still gets a very closed makespan value of 1898.0. On tardiness, constrained learning agents show a lower value than the others in most instances. On Instance01, Instance04 and Instance08, SAC agent gets the lowest tardiness. However, it is notable that we care more about constraint satisfaction rather than minimising tardiness. Even though in Instance04, SAC gets the lowest tardiness value with 28.9, we consider RCPOM as the best agent since it gets the lowest makespan value of 1956.9, whose tardiness is under the constraint threshold, i.e., $40.5<50$. 
A similar case is also observed in unseen instances, demonstrated in Tab. \ref{tab:unseenResult}. Learning agents still perform better on unseen instances compared with dispatching rules, except STD which gets the best result on Instance16 with constraint satisfaction. RCPOM, the proposed method, achieves the best average makespan and is statistically better than almost other policies on Instance10-14. 



\subsection{Discussion}


\subsubsection{Mediocre dispatching rules}

\renewcommand{\arraystretch}{1}

\begin{table*}[h]
\normalsize
    \centering
        \caption{Average makespan and tardiness over 30 independent trails on training set (Instance01-08). The bold number indicates the best makespan and tardiness.``+/$\approx$/-'' indicates the agent performs statistically better/similar/worse than RCPOM agent. ``M/C (P)'' indicates the average normalised makespan, tardiness and percentage of constraint satisfaction. The number of policies that RCPOM is better than others in terms of makespan and tardiness on each instance is given in the bottom row. }
    \resizebox{\textwidth}{!}{
      \setlength{\tabcolsep}{1pt}
    \begin{tabular}{c|c|c|c|c|c|c|c|c|c}
\toprule
\multicolumn{1}{c|}{\multirow{2}{*}{Algorithm} }&Instance01 & Instance02 & Instance03 & Instance04 & Instance05 & Instance06 & Instance07 & Instance08 & \multirow{2}{*}{M/C (P)} \\
&$F_m$/$F_t$ & $F_m$/$F_t$& $F_m$/$F_t$& $F_m$/$F_t$& $F_m$/$F_t$& $F_m$/$F_t$& $F_m$/$F_t$& $F_m$/$F_t$&\\
\midrule
RCPOM & 1840.0/56.7 & \textbf{1908.4}/48.2 & 1914.8/45.0 & \textbf{1956.9}/40.5 & \textbf{1852.5}/\textbf{21.3} & \textbf{1911.1}/\textbf{41.0} & \textbf{1927.0}/\textbf{8.4} & 1898.0/21.0&\textbf{0.97}/0.89 (75\%)\\
RCPOM-NS & 1891.9-/58.8$\approx$ & 1979.5-/51.6$\approx$ & 1931.1$\approx$/46.9$\approx$ & 2003.9-/53.0- & 2028.8-/58.1- & 2028.9-/88.8- & 1951.4-/15.2- & 1951.0-/41.8-&0.68/0.70 (56\%)\\
IPO & 1860.5-/48.6+ & 1933.5$\approx$/50.4$\approx$ & 1924.9$\approx$/47.6$\approx$ & 1979.1-/49.6$\approx$ & 1951.1-/47.0- & 1951.5-/71.4- & 1977.0-/14.8- & 1943.0-/34.9-&0.80/0.77 (61\%)\\
L-SAC & 1884.8-/60.3$\approx$ & 1987.4-/57.2- & 1933.7$\approx$/44.0$\approx$ & 1973.5$\approx$/42.2$\approx$ & 1955.4-/44.7- & 1992.9-/68.5- & 1983.9-/19.8- & 1923.6-/32.8-&0.74/0.76 (58\%)\\
L-PPO & 1872.1-/59.5$\approx$ & 1963.5-/52.3$\approx$ & \textbf{1910.3}$\approx$/49.7$\approx$ & 1984.0-/49.7- & 1954.0-/55.5- & 1968.6-/75.6- & 1979.7-/18.9- & 1899.2$\approx$/32.6-&0.80/0.73 (53\%)\\
SAC & \textbf{1798.4}+/49.1+ & 1950.0-/68.9- & 1965.0-/53.4- & 2002.0-/\textbf{28.9}+ & 1900.0-/26.6- & 2011.9-/71.3- & 1927.0$\approx$/12.6- & 1914.1-/\textbf{15.7}+&0.82/0.83 (62\%)\\
PPO & 1858.0$\approx$/52.3+ & 1941.8$\approx$/49.9$\approx$ & 1918.2$\approx$/44.3$\approx$ & 1995.3-/55.6- & 1937.1-/47.6- & 1961.0-/82.0- & 1988.6-/14.6- & 1920.7$\approx$/35.6-&0.79/0.75 (61\%)\\
\midrule
FCFS & 2081.1-/90.2- & 2084.5-/82.3- & 2014.7-/64.2- & 2136.7-/105.0- & 2194.6-/122.5- & 2123.5-/85.2- & 1927.9$\approx$/11.2- & 1933.6-/27.4$\approx$&0.31/0.48 (37\%)\\
EDD & 1903.2$\approx$/\textbf{33.9}+ & 1968.6-/\textbf{29.6}+ & 1977.8-/\textbf{26.5}+ & 1988.1-/32.8+ & 1950.7-/33.7- & 2016.8-/46.8- & 1940.5-/12.1- & 2020.8-/45.3$\approx$&0.66/\textbf{0.91} (\textbf{86\%})\\
NVF & 1876.5$\approx$/69.6$\approx$ & 1958.4-/56.4$\approx$ & 1946.7-/49.6$\approx$ & 2040.6-/66.7- & 1933.9-/56.3- & 1953.4-/78.5- & 1996.5-/21.8- & 1944.6-/35.5-&0.71/0.67 (51\%)\\
STD & 1868.8$\approx$/66.6$\approx$ & 1961.7-/49.2$\approx$ & 1921.3$\approx$/51.7- & 1970.7-/35.6+ & 1917.1-/41.4- & 1955.4-/62.7- & 1983.7-/30.4- & \textbf{1883.5}+/35.1-&0.83/0.75 (53\%)\\
Random & 2098.7-/124.5- & 2113.8-/103.1- & 2091.2-/143.7- & 2135.1-/123.1- & 2149.3-/119.0- & 2159.1-/129.3- & 2083.0-/70.2- & 2067.8-/89.5-&0.02/0.00 (12\%)\\

\midrule
 & 6/2 &9/4 &5/4 &10/6 &11/11 &11/11 &9/11 &8/8 \\

\bottomrule
    \end{tabular}}

    \label{tab:result}
\end{table*}

\begin{table*}[h]
\normalsize
    \centering
        \caption{Average makespan and tardiness over 30 independent trails on test set (Instance09-16). The Bold number indicates the best makespan and tardiness.``+/$\approx$/-'' indicates the agent performs statistically better/similar/worse than RCPOM agent. ``M/C (P)'' indicates the average normalised makespan, tardiness and percentage of constraint satisfaction. The number of policies that RCPOM is better than others in terms of makespan and tardiness on each instance is given in the bottom row. }
    \resizebox{\textwidth}{!}{
      \setlength{\tabcolsep}{1pt}
    \begin{tabular}{c|c|c|c|c|c|c|c|c|c}
\toprule
\multicolumn{1}{c|}{\multirow{2}{*}{Algorithm} }&Instance09 & Instance10 & Instance11 & Instance12 & Instance13 & Instance14 & Instance15 & Instance16 & \multirow{2}{*}{M/C (P)} \\
&$F_m$/$F_t$ & $F_m$/$F_t$& $F_m$/$F_t$& $F_m$/$F_t$& $F_m$/$F_t$& $F_m$/$F_t$& $F_m$/$F_t$& $F_m$/$F_t$&\\
\midrule

RCPOM & 1840.0/57.1 & \textbf{1917.6}/49.4 & \textbf{1900.3}/45.8 & \textbf{1954.3}/38.8 & \textbf{1862.7}/\textbf{18.4} & \textbf{1914.1}/\textbf{41.8} & 1960.8/19.6 & 1896.4/\textbf{21.3}&\textbf{0.95}/\textbf{0.90} (\textbf{74\%})\\
RCPOM-NS & 1904.6-/60.2$\approx$ & 1985.4-/52.0$\approx$ & 1931.6-/47.3$\approx$ & 1995.6-/51.0- & 2009.1-/55.0- & 2034.9-/92.6- & 1956.6$\approx$/16.2+ & 1941.0$\approx$/40.3-&0.68/0.72 (55\%)\\
IPO & 1860.5-/\textbf{48.6}+ & 1933.5$\approx$/50.4$\approx$ & 1924.9-/47.6$\approx$ & 1979.1-/49.6- & 1951.1-/47.0- & 1951.5-/71.4- & 1977.0-/14.8+ & 1943.0-/34.9-&0.80/0.80 (61\%)\\
L-SAC & 1892.3-/62.0$\approx$ & 1983.6-/56.9$\approx$ & 1933.1-/43.4+ & 1974.5-/43.0$\approx$ & 1954.4-/41.0- & 1999.4-/70.9- & 1981.5-/18.6+ & 1930.3-/34.8-&0.73/0.78 (58\%)\\
L-PPO & 1877.5-/60.8$\approx$ & 1961.3-/52.6$\approx$ & 1905.8$\approx$/49.5$\approx$ & 1980.9-/49.8- & 1960.2-/50.1- & 1965.5-/78.5- & 1975.1-/18.2+ & 1901.5$\approx$/33.1-&0.81/0.76 (57\%)\\
SAC & \textbf{1801.3}+/49.3+ & 1950.0-/69.1- & 1956.0-/56.8- & 2000.0-/\textbf{28.6}+ & 1904.0-/27.3- & 2015.1-/71.1- & \textbf{1930.0}+/12.6+ & 1987.0-/38.2-&0.76/0.81 (62\%)\\
PPO & 1863.1-/53.5+ & 1941.8$\approx$/50.1$\approx$ & 1913.2$\approx$/44.1$\approx$ & 1986.3-/55.0- & 1927.4-/41.4- & 1965.5-/85.9- & 1990.1-/15.0+ & 1908.2$\approx$/35.4-&0.81/0.77 (62\%)\\
\midrule
FCFS & 2089.3-/92.5- & 2045.4-/75.7- & 1996.9-/67.6- & 2107.1-/95.2- & 2191.9-/121.7- & 2130.1-/89.9- & 1934.1$\approx$/11.1+ & 1946.8-/35.8-&0.32/0.48 (35\%)\\
EDD & 1996.9-/107.7- & 1976.3-/\textbf{33.4}+ & 1978.7-/\textbf{22.0}+ & 1997.6-/36.9$\approx$ & 1962.1-/37.0- & 1993.5-/44.6- & 1934.8$\approx$/\textbf{11.0}+ & 2052.3-/69.4-&0.59/0.79 (73\%)\\
NVF & 1847.5$\approx$/57.2- & 1958.8-/54.3$\approx$ & 1926.6$\approx$/51.8$\approx$ & 2021.7-/65.9- & 1939.2-/41.2- & 1975.5-/89.8- & 2003.5-/19.3+ & 1933.8-/40.9-&0.73/0.71 (58\%)\\
STD & 1894.1-/72.5- & 1958.2$\approx$/49.7$\approx$ & 1923.1$\approx$/52.4- & 1985.2-/38.1$\approx$ & 1905.1-/31.4- & 1974.3-/71.9- & 1990.3$\approx$/33.0- & \textbf{1885.2}+/38.6-&0.80/0.76 (55\%)\\
Random & 2132.7-/135.3- & 2100.9-/99.5- & 2076.2-/144.4- & 2097.5-/106.3- & 2144.1-/102.6- & 2142.9-/123.6- & 2101.3-/81.1- & 2055.1-/94.1-&0.03/0.02 (11\%)\\
\midrule
&9/5 &8/3 &7/4 &11/7 &11/11 &11/11 &6/2 &7/11 \\

\bottomrule
    \end{tabular}
    }

    \label{tab:unseenResult}
\end{table*}

Dispatching rules perform promisingly on 16 instances, as shown in Tab. \ref{tab:result} and Tab. \ref{tab:unseenResult}. EDD usually has the lowest tardiness and STD even gets the lowest makespan on Instance08 and Instance16.
 FCFS and EDD are time-related type dispatching rules. FCFS always chooses tasks according to their arrival time, i.e., assigns the task that arrives first. EDD will select the task that has the earliest due date. In contrast to time-related rules, NVF and STD are distance-related rules. They both optimise the objective that strongly relates to distance such as makespan. The difference is that NVF selects the task with the nearest pickup point while STD selects the task with the shortest distance from the current position to the pickup point and then the delivery point.

Dispatching rules use simple manual mechanisms to schedule tasks and achieve a promising performance, which are usually better than the Random agent. However, such mechanisms may not be able to handle more complex scenarios. And it is hard to improve the dispatching rule due to its poor adaptability. From Fig. \ref{fig:box}, it is clear that learning agents usually perform better than the dispatching rules with lower averaged makespan on the instances. Dispatching rules keep using the same mechanism in the long sequential decision process. It makes sense that they show a limited performance since one rule usually works in certain specific situations.
Instantaneous modifications have to be made when unexpected events occur. 
For example, EDD and FCFS consider time-related objectives such as makespan which can be easily determined in Eq. \ref{eq:tardiness}. Both rules optimise the objective partially. When the instance has a large concession space for delayed tasks, EDD and FCFS may hardly work. In the problem considered in this paper, two objectives (i.e., distance and time) are involved. Although tardiness is considered a constraint, it is hard to identify the correlation between makespan and tardiness. 
It is shown on Instance08 that NVF and STD are better in both makespan and tardiness. But it can also be seen that in Instance01 and Instance03, EDD has lower tardiness, whose makespan is worse than NVF and STD. 
This observation implies that the ability of simple dispatching rules is limited. Our hybrid action space is motivated by the phenomenon and expected to provide more optimisation possibilities and adaptability.

\def\nouse{
\subsubsection{Benefit from CMDP}

From Fig. \ref{fig:box}, it is clear that learning agents usually perform better than the dispatching rules with lower averaged makespan on the instances. Dispatching rules keep using the same mechanism in the long sequential decision process. They show limited performance since one rule usually works in certain specific situations. The usage of the hybrid action space provides more optimisation possibilities. 
}

\subsubsection{Constraint handling helps}
\begin{figure}[t]
    \centering
    \subfigure{
		\centering
		\includegraphics[width=0.83\columnwidth]{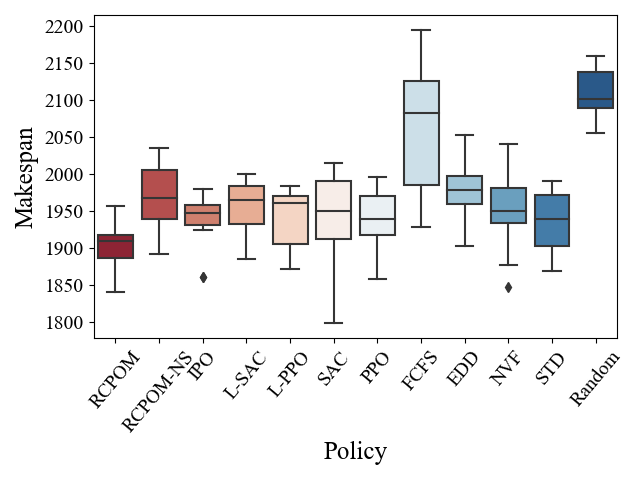}
  
  }	

     \subfigure{
		\centering
		\includegraphics[width=0.83\columnwidth]{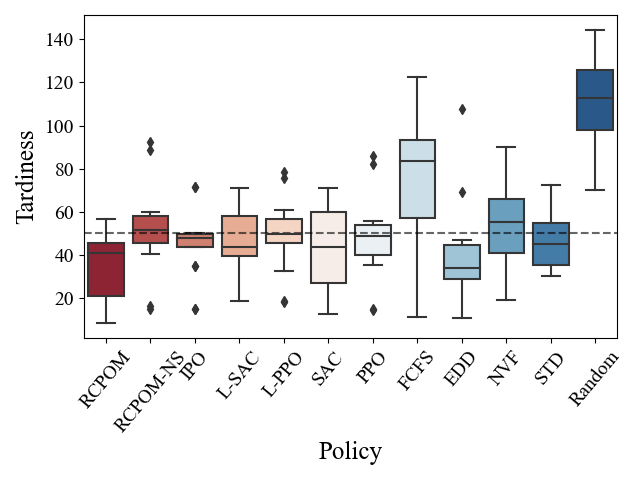}
	}
	\caption{\label{fig:box}Performance averaged over 16 instances.}
\end{figure}


A vehicle should be available when assigning a task, which is critical for safe manufacturing. Previous work~\cite{hu2020deep} resamples an action in case of constraint violations, which does not work out in our scenario. Tab. \ref{tab:ptime} shows the time cost for determining task assignments. The agents using the invalid action masking technique take much shorter decision-making time in contrast to the timeout by~\cite{hu2020deep}. The application of the invalid action masking technique guarantees instantaneous safety. Sampling once is enough to obtain a valid action. Another benefit of masking is the compression of action space. The agent can focus more on the choice of valid actions and further improve the exploration efficiency.

Even though SAC and PPO agents show remarkable performance in terms of makespan, they fail in tardiness. SAC agent gets a high tardiness value on Instances indexed with 2, 3, 6, 10, 11 and 14. It is attributed to the single-attribute reward function since it only relies on the makespan and does not consider tardiness. A straightforward method to handle the cumulative constraint is augmenting the objective function. L-SAC and L-PPO reshape the reward function with a fixed Lagrangian multiplier, seeing Eq. \eqref{eq:reshapeR}. However, it is hard to decide the multiplier. From Tab. \ref{tab:result}, L-SAC and L-PPO do not outperform SAC and PPO much, even L-PPO has the lowest makespan on Instance03. RCPOM provides a more flexible way to restrict the behaviour of policy with a suitable multiplier that is updated during training.
It is demonstrated in Tab. \ref{tab:result}, Tab. \ref{tab:unseenResult} and Fig. \ref{fig:box}, RCPOM outperforms other learning agents on average. RCPOM has a slight gap with EDD on tardiness but outperforms it much on makespan.
Although IPO is also a CRL algorithm, its assumption that the policy should satisfy constraints upon initialisation~\cite{liu2021policy} limits its performance when solving the DMH problem.

\subsubsection{Promising performance on unseen instances}

To further validate the performance of the proposed method, we also test it on some unseen instances from Instance09 to Instance16. These unseen instances are generated by mutating the training instances. Our method still outperforms others on the unseen instances according to Tab. \ref{tab:unseenResult}. Such a stable performance gives more possibility to apply our method to real-world problems, meeting dynamic and complex situations.

\subsubsection{Invariant reward shaping improves}

The raw reward function of the process is the negative makespan, which is rewarded at the end, denoted in Eq. \eqref{eq:rewardf}. It is challenging for an RL agent to learn such a sparse reward function with only negative values. The consequence brought by the lack of positive feedback is that agents may be stagnant and conservative. In almost all instances, RCPOM agent with reward shaping performs better than the one without reward shaping, shown in Tab. \ref{tab:result} and Tab. \ref{tab:unseenResult}. Although the relative reward value is modified by the reward shaping, it is proved that the optimal policy keeps invariant. Invariant reward shaping helps agents explore more and make the most of the positive feedback.



\section{Conclusion}
\label{sec:conclusion}
This paper studies the dynamic material handling problem. Newly arrived tasks and vehicle breakdowns are considered dynamic events. Due to the lack of free simulators and problem instances, we develop a gym-like simulator, namely DMH-GYM and provide a dataset of diverse instances. Considering the constraints of tardiness and vehicle availability, we formulate the problem as a constrained Markov decision process. A constrained reinforcement learning algorithm that combines Lagrangian relaxation and invalid action masking, named RCPOM is proposed to meet the hybrid cumulative and instantaneous constraints. We validate the outstanding performance of our approach on both training and unseen instances. The experimental results show RCPOM statistically outperforms state-of-the-art reinforcement learning agents, constrained reinforcement learning agents, and several commonly used dispatching rules. It is also validated that the invariant reward shaping helps. The effectiveness of RCPOM provides a new perspective for dealing with real-world scheduling problems with constraints. Instead of constructing a complicated reward function manually, it is possible to restrict the policy directly by constrained reinforcement learning methods.

As future work, we are interested in extending the problem by introducing more dynamic events and constraints that widely exist in real-world scenarios.

\balance
\bibliographystyle{IEEEtran}
\bibliography{main}

\end{document}